\documentclass[10pt,twocolumn]{article}

\pdfoutput=1

\usepackage{times}
\usepackage{epsfig}
\usepackage{graphicx}
\usepackage{amsmath}
\usepackage{amssymb}
\usepackage{subfigure}
\usepackage{xspace}
\usepackage{units}
\usepackage[usenames]{color} 
\usepackage{algorithm}
\usepackage{algorithmic}

\usepackage[pagebackref=true,breaklinks=true,colorlinks,bookmarks=false]{hyperref}

\def\etal{et al.\@\xspace}

\graphicspath{{./Figures/}}

\newcommand{\widthfrac}{0.9}

\definecolor{shadecolour}{gray}{0.4}

\newcommand{\mat}[1]{{\mathbf #1}} 
\renewcommand{\vec}[1]{{\mathbf #1}} 

\newcommand{\est}[1]{{\widehat{#1}}}

\DeclareMathOperator*{\argmin}{arg\,min}
\definecolor{shadecolour}{gray}{0.4}

\newcommand{\Rdim}{\mathbb{R}}
\newcommand{\Bdim}{\mathbb{Z}_2}

\newcommand{\nTemplates}{T}
\newcommand{\template}{\vec{T}}
\newcommand{\tIndex}{t}

\newcommand{\y}{\vec{y}}
\newcommand{\proj}{\mat{P}}
\newcommand{\nImages}{P}
\newcommand{\projIndex}{p}
\newcommand{\synthImage}{{\mat{I}}}
\newcommand{\tSil}{{\mat{\pi}}} 
\newcommand{\MSil}{\y}
\newcommand{\SilLength}{S}
\newcommand{\nMeasurements}{D}

\newcommand{\Hash}{\boldsymbol{\Phi}}
\newcommand{\PhiDensity}{k}
\newcommand{\Basis}{\boldsymbol{\Pi}}
\newcommand{\HBasis}{\boldsymbol{\Psi}}

\newcommand{\coeffs}{{\boldsymbol\alpha}}

\newcommand{\lam}{\lambda}

\newcommand{\lzero}{{\ell}^0}
\newcommand{\lone}{{\ell}^1}

\def\eg{eg.\@\xspace}

\newcommand{\Hashk}{m}
\newcommand{\Hashn}{n}

\newcommand{\shortcite}{\cite}

\begin{document}

\title{Deconstruction of compound objects from image sets}

\author{Anton~van~den~Hengel, John~Bastian, Anthony~Dick, Lachlan~Fleming\\
The Australian Centre for Visual Technologies\\
The University of Adelaide\\
{\tt\small Anton.vandenHengel@adelaide.edu.au},
{\small\url{http://acvt.com.au}}
}

\maketitle

\begin{abstract}
%
%

We propose a method to recover the structure of a compound object from multiple silhouettes. Structure is expressed as a collection of 3D primitives chosen from a pre-defined library, each with an associated pose. This has several advantages over a volume or mesh representation both for estimation and the utility of the recovered model. The main challenge in recovering such a model is the combinatorial number of possible arrangements of parts. We address this issue by exploiting the sparse nature of the problem, and show that our method scales to objects constructed from large libraries of parts.

\end{abstract}


\section{Introduction}
\label{sec:Inro}

We propose a method to estimate the structure of compound objects from a set of images.
Such objects are prevalent in our everyday environment, and in many cases our knowledge of their innate structure is essential to our understanding of them. They include man made objects such as buildings, furniture, and cars, but also natural objects such as trees and plants. 
Our goal is to find the simplest construction which explains the shape of the object, using a given library of parts.  Unlike most work on the recovery of shape from images, our method does not generate a point cloud, or a volume, but a structural explanation of way the object depicted is constructed. In this sense it is aligned with the blocks-world approach~\cite{roberts65}, recently revisited by~\cite{gupta10blocks}. 

\begin{figure}[h]
	\centering
   \includegraphics[height=1.2in]{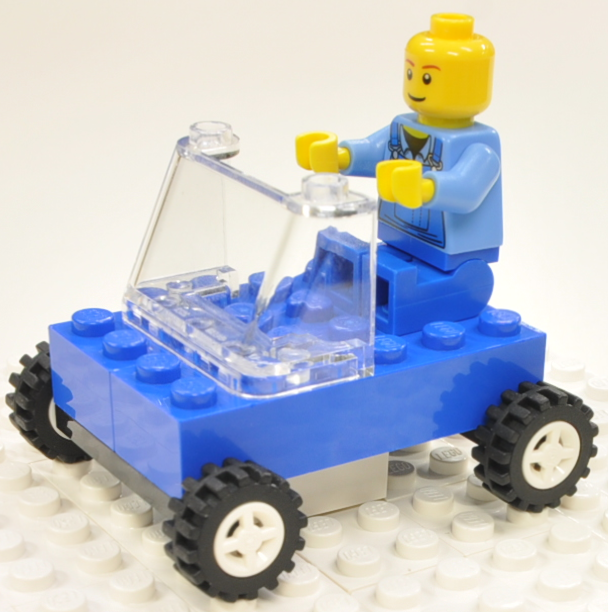}
 \quad   \includegraphics[height=1.2in]{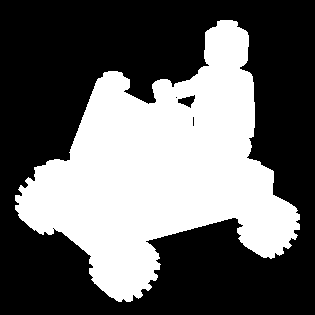}
   \includegraphics[height=1.2in]{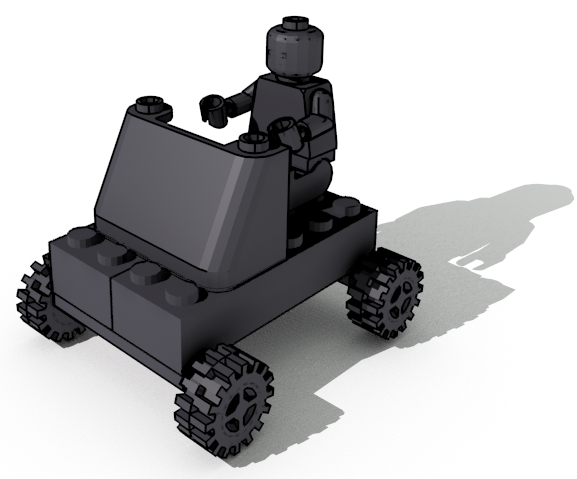}
   \includegraphics[height=1.2in]{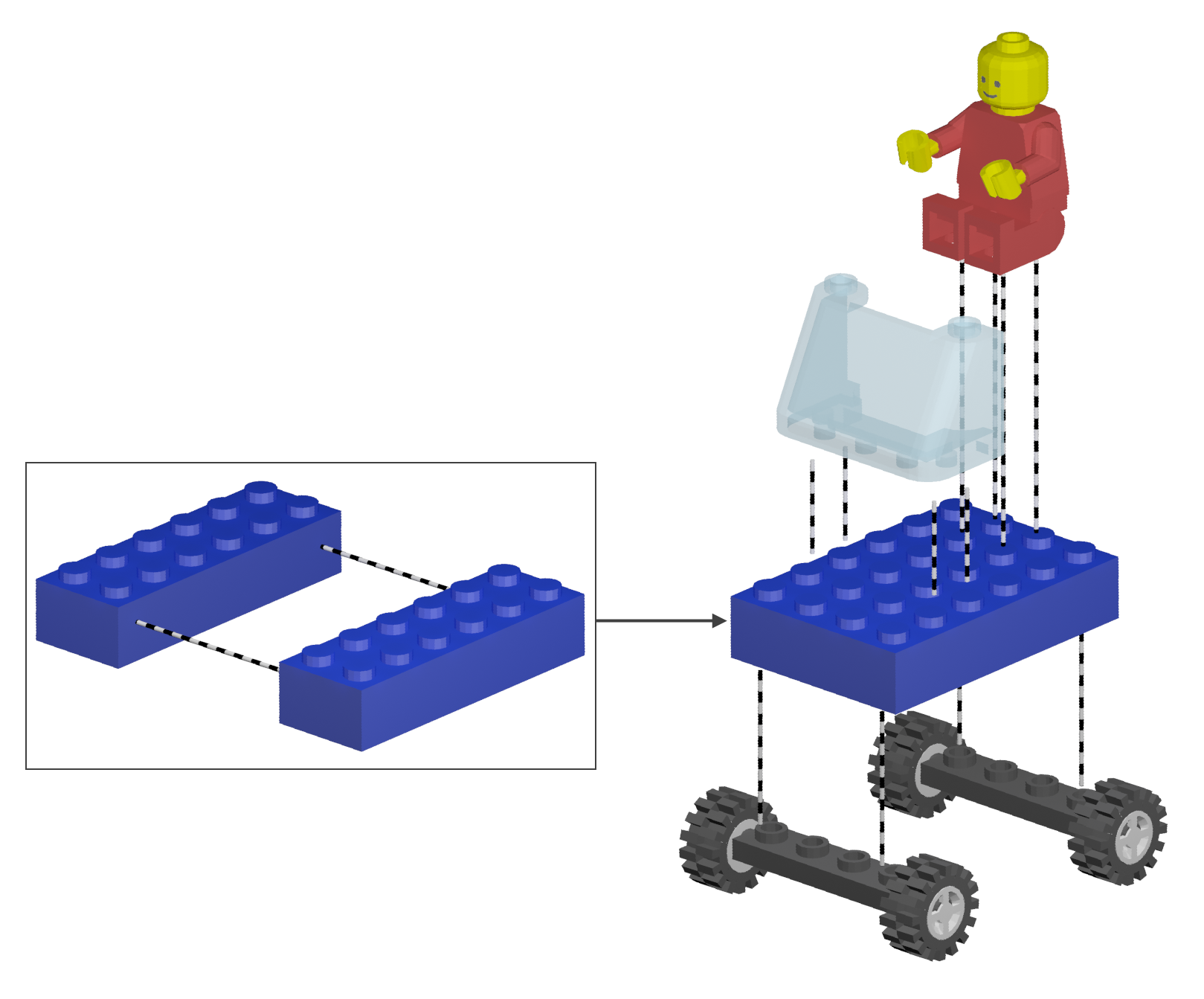}
   \caption{An illustration of the deconstruction process, from image set, and silhouettes, to an estimate of the building blocks from which an object is constructed, and how they fit together.}
   \label{fig:teaser}
\end{figure}

The method we propose 
reasons in 3D about the structure of an object on the basis of its appearance in an image set.
%
This requires a set of building blocks from which an object might be composed.  As we are interested in structure, rather than appearance, these building blocks are described in terms of their shape and position only.  The recovered structure estimate is the smallest set of building blocks required to reconstruct the object in question.

By estimating structure, rather than shape, the method provides a semantic interpretation of the elements from which the object is constructed.  This means that the resulting structure estimate can be used to animate the object, decompose it, or augment it.
For example, the estimated structure of a car 
identifies the doors and wheels, as is required to rig the object in order to animate each component appropriately.  This in turn allows the modelled object to be incorporated into a video game or video sequence without further work.  
Our method thus differs significantly from classical Structure from Motion methods; we are effectively deconstructing the object into its constituent parts.

Representing an object by 3D primitives also has advantages for the estimation of its shape. The space of possible primitive arrangements is far smaller than the space of possible combinations of 3D vertices or voxels, resolving ambiguities and eliminating shapes that are impossible. 
However, the solution space remains extremely large for complex objects, and one of our main contributions in this paper is a way of formulating the primitives and optimising their arrangement so that a global optimum, corresponding to the simplest configuration that explains the input silhouettes, can be found within this space.

After discussing related work in Section~\ref{sec:Related}, we define the problem and describe the form of the measurements and the 3D building blocks in Section~\ref{sec:Method}. In Section~\ref{sec:Optimisation}, we describe how we solve the problem in practice, followed by some experimental results in Section~\ref{sec:Results}.

Our notation is as follows.
We label scalars in non-bold italic typeface, with lowercase indicating an index and upper case its limit, thus $\projIndex \in \{1,\ldots,\nImages\}$.  Vectors are represented as bold 
lower case letters (eg. $\mat{x}$), and are column vectors.  Matrices appear as bold 
upper case letters (\eg $\mat{X}$).  

\begin{figure}[ht]
	\centering
		\includegraphics[width=1.00\columnwidth]{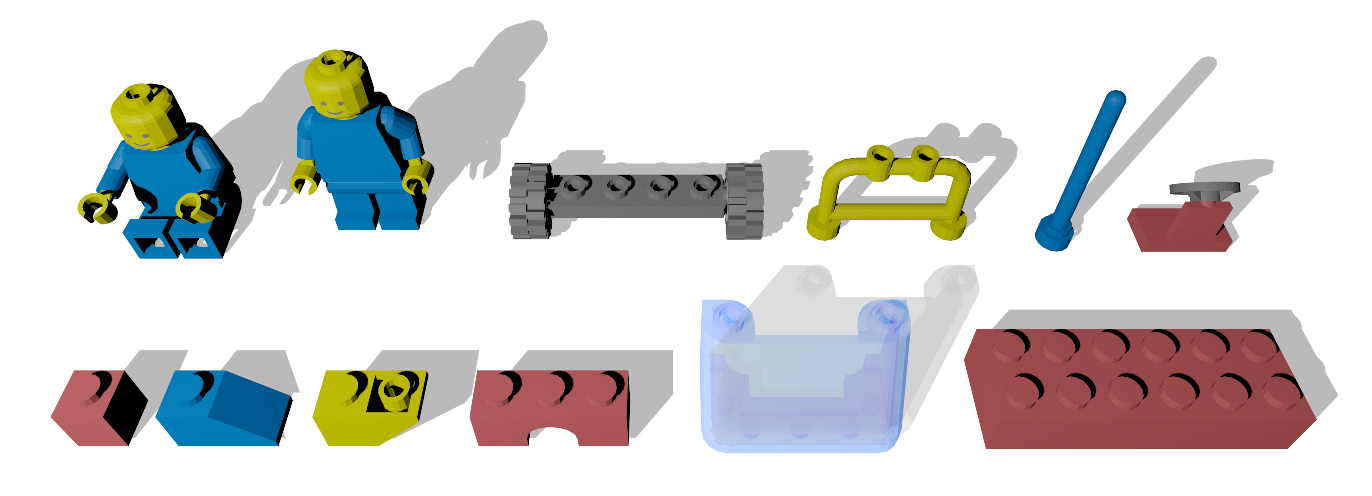}
		\caption{A selection of the shape models used in recovering the structure of the Lego\textsuperscript{\textregistered} models.  Each such shape model must be rendered in every location and orientation in which is may appear to generate the set of templates.}
	\label{fig:LegoPieces}
\end{figure}

%
%


\subsection{Related Work}
\label{sec:Related}

The idea of analysing the composition of a visible object or scene in terms of a set of basis parts has a long history in computer vision, stretching back to the blocks-world interpretation of flat shaded polyhedral surfaces in the 1960s~\cite{roberts65}.
This was further developed to interpret collections of simple shapes~\cite{brooks1979acronym}, to reason about structure from physical constraints~\cite{brand95}, and, recently, to inform the recovery of a qualitative scene reconstruction from a single image~\cite{gupta10blocks}. 
There are also a variety of methods which interpret images or point clouds using predefined families of surfaces, or implicit functions  (see for example \cite{Taubin:1991}, and for a reverse engineering application~\cite{Fitzgibbon97high-levelcad}).  Paramterised sets of transformations have also been used to identify the building blocks of building facades\cite{eth_biwi_00530} and 3D structures\cite{pmwpg_structure_sig_08}.

Our approach, however, poses structure recovery as a classification problem where evidence is sought within the image set for the existence of each of the possible building blocks from which the object might be constructed.  
This approach is thus more closely related to work such as 
\cite{Ullman91} which describes a method for object recognition using linear combinations of images.  They recognise objects by reconstructing the location of features in a query image as a linear combination of those in images of database objects.  
However, the method aims only to recognise objects for which it has a pre-existing model.

Wright \etal in~\cite{Wright08robustface}  build on the eigenface-based face recognition method of of Turk and Pentland~\shortcite{Turk91} to identify a query image of a face by analysis of the coefficients of its reconstruction from a set of basis vectors.  They form the basis vectors by random projection of the training images, and apply the tools of sparse representation and compressed sensing to the problem.  The reconstruction of the query image from the basis set is defined as that miminising the $\lzero$ norm and is under-constrained, but the presumed sparsity in the space of face images allows a solution using the $\lone$ norm in its place.  This method is similar to that proposed here in that it uses the coefficients which best (linearly) reconstruct the data for the purposes of classification. 

Shakhnarovich \etal in~\cite{Shak03} propose Parameter Sensitive Hashing (PSH) as a method for estimating human pose from a single silhouette.  The method thus recovers pose from a projection, as we aim to do.  
A similar problem is tackled in~\cite{grauman03} using a Bayesian human shape model.
%
%
These methods aim to recover the pose of a single predefined structure, however, rather than recovering pose and structure simultaneously.

Our method also has similarities to the non-rigid factorisation approach proposed by Bregler \etal~\cite{Bregler00} whereby a the shape of a deforming object is recovered as a linear
combination of basis shapes.  Although a number of improvements have been proposed (see  \cite{Paladini09} for example), the method fundamentally operates on identified sets of feature points on the object, and aims to recover shape rather than structure.  These methods also typically require a very specific camera model and assume that the structure linking the parts together is fixed and known.


\section{Estimating structure}
\label{sec:Method}

Given one or more measured images, flattened and concatenated into a $\SilLength$ dimensional vector $\y$, and a collection of $\nTemplates$ template vectors, stacked into a $\SilLength \times \nTemplates$ matrix $\Basis$, we seek the most parsimonious combination of templates that reconstructs the observed image data:
\begin{align}
	 \argmin_\coeffs \|\coeffs \|_0, \mbox{ s.t. } \y = \Basis \coeffs \label{eq:1}
\end{align}
where $\coeffs$ is a length $\nTemplates$ vector.
Due to practical considerations, we adopt a relaxed formulation in later sections of the paper, but
%
what differentiates our approach from others based on linear combinations of building blocks, is the form of the measurements, the nature of the building blocks, and the process by which the coefficients are estimated.

\subsection{The measurement vector}
\label{sec:Measurements}

Because the template shapes do not have any associated appearance information, we 
convert each image to a silhouette. 
Each pixel in each silhouette image records the presence or absence of the object at a particular location in one of the images as a 1 or a 0 respectively.  




We consider multiple silhouette images of an object, each of which has a corresponding  projection matrix $\proj^\projIndex$ (where $\projIndex \in \{1,\ldots,\nImages\}$) as in Figure~\ref{fig:lego_minfig_silhouettes}.  We assume that these projection matrices are available a priori, or calculable from the image set.  In the case of the experiments presented in Section~\ref{sec:Results}, the $\proj^\projIndex$ were estimated from the image sets using standard camera calibration software.

The silhouette of the object in each image is flattened into a vector $\MSil^{\projIndex}$.  We then form the $\SilLength$ dimensional binary vector $\MSil$ by concatenating the $\MSil^{\projIndex}$ for all $\projIndex$.  

%

The advantage of silhouettes as a cue is that they are dependent on shape alone.  There is no reliance on texture to generate identifiable feature points, on an active sensor to generate a point cloud, or the interaction with a light source to generate usable shading variations.  Most importantly, however, the silhouettes of components can be composited in order to form the silhouette of the whole.

\subsection{The template shapes}
\label{sec:templates}

Each template is defined by a 3D shape and a 3D pose. 
These templates are the elements from which each shape model is constructed.  
In the case of the Lego example each template shape represents a particular type of block at a specific position and orientation, a selection of which are shown in Figure~\ref{fig:LegoPieces}.  The set of templates includes every shape that might form part of an object to be reconstructed, at every location it can occur.




\begin{figure}[h]
	\centering
		\includegraphics[width=1.00\columnwidth]{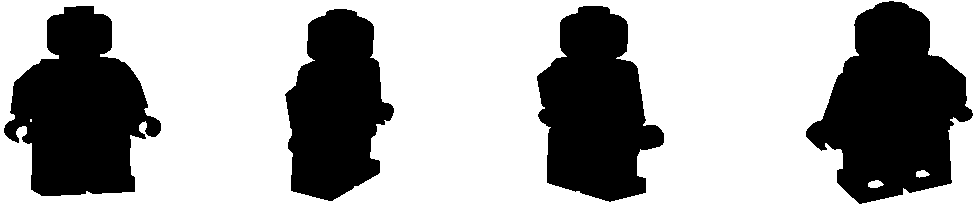}
	\caption{The silhouettes of one template, which are flattened and concatenated 
	to form a silhouette vector.}
	\label{fig:lego_minfig_silhouettes}
\end{figure}

\begin{figure*}[ht]
	\centering
		\includegraphics[width=1.00\textwidth]{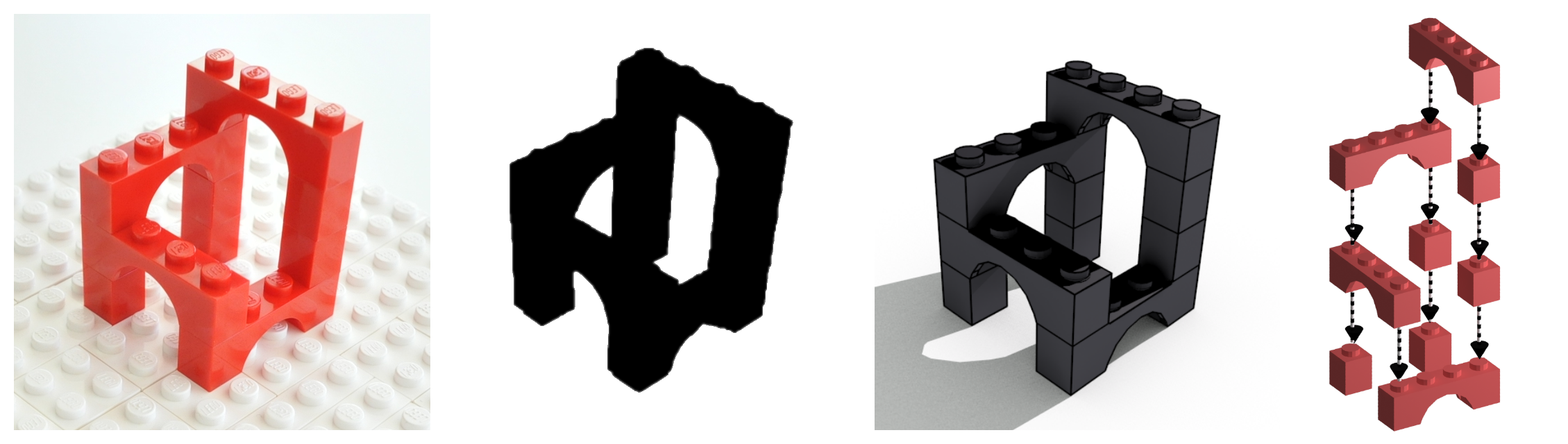}
	\caption{Lego Escher arches: estimating structure, rather than just shape, provides semantic information such as how objects are constructed.  In this example the structure estimate calculated on the basis of of 4 silhouettes allows an analysis of the components of the object and how they fit together.}
	\label{fig:lego_teaser}
\end{figure*}


Generating the basis matrix $\Basis$ 
requires calculating a set of silhouettes for each template shape, such as those in Figure~\ref{fig:lego_minfig_silhouettes}.
Each 
template $\template_\tIndex$ is rendered using each projection matrix $\proj^\projIndex$ to produce a corresponding synthetic image 
$\synthImage_{\tIndex}^{\projIndex}$.  The silhouette of each of these synthetic images is
then flattened into a vector.  The silhouette vectors for each template $\template_{\tIndex}$ are  then concatenated into a single length $\SilLength$ vector 
$\tSil_{\tIndex}$.
There are $\nTemplates$ such  vectors $\tSil_{\tIndex}$, and these make up the columns of the  matrix $\Basis \in \Bdim^{\SilLength \times \nTemplates}$.  The columns of matrix $\Basis$ thus represent a basis of template silhouettes from which we aim to construct the true object silhouette according to Equation~\eqref{eq:1}.
Each silhouette image $\synthImage_{\tIndex}^{\projIndex}$ depends on both $\template_\tIndex$ and $\proj^\projIndex$, which  means that both must be known before $\tSil_{\tIndex}$ can be formed.  

An advantage of this method is that we need only render each component individually, rather than in every possible combination. This means that the number of templates scales approximately linearly with the number of components rather than combinatorially with the complexity of the object to be modelled.


We demonstrate the technique on Lego models as an indication of a scenario which is an intuitive, but challenging, application.  Lego models are constructed of many small parts, each of which may appear in hundreds of different positions.  This is far in excess of the number and variety of components that we would expect normal objects to be composed from.  We show below that structure estimation can be posed as a linear programming problem, and thus that tens of thousands of template shapes can be analysed if modern linear programming techniques are used.




\section{Structure recovery as linear programming}
\label{sec:Optimisation}

%
%
%


In order to solve~\eqref{eq:1}, we replace the $\lzero$ norm on $\coeffs$ with the $\lone$ norm, thus rendering the problem convex.  Donoho and Huo in~\shortcite{Donoho01} show that, under certain sparsity conditions, the relaxed equivalent of~\eqref{eq:1}:
\begin{align}
	 \argmin_\coeffs \|\coeffs \|_1, \mbox{ s.t. } \y = \Basis \coeffs \label{eq:2}
\end{align}
has the same solution.
%
%
The sparsity constraint requires that each object to be reconstructed must be expressible using a very small fraction of the large set of possible templates, and therefore   
that we expect the fraction of non-zero elements in $\coeffs$ to be very small.  The length of $\coeffs$ depends upon the number of template shapes required to achieve a reasonable quality reconstruction, but is practically limited by the scale of optimisation problem which we can  solve.  
Modern Linear Programming packages are capable of solving systems with millions of variables, although the largest problem solved in the process of generating the results below had over $72,000$ templates.
The objects we seek to analyse are typically constructed of tens of template shapes so the vector $\coeffs$ can be expected to have approximately one thousandth of its elements not equal to zero. This level of sparsity is more than enough to satisfy the requirements of Donoho \etal~\shortcite{Donoho01}.


%
 
Due to noise in the silhouette images, and the fact that no real set of templates is likely to perfectly explain the shape of all suitable objects, the equality constraint in~\eqref{eq:2} is unattainable in practical situations.  
A number of relaxations of such a constraint have been proposed in the literature, but our concern is whether a combination of template silhouettes are as close as possible to the silhouette of the original object.  This would suggest the $\lzero$ norm, but as previously described practicality demands the $\lone$ norm instead.  We thus seek to optimize
\begin{equation}
\argmin_\coeffs \|\y - \Basis \coeffs \|_1 + \lam \|\coeffs\|_1, \mbox{ s.t. } 0 \leq \coeffs \leq 1,
\label{eq:Relaxed}
\end{equation}
where $\lam$ is a scale factor controlling the degree to which the method focuses on reconstruction error or parsimony.
%
The bounds on values of the elements of $\coeffs$ ensure that the silhouette of the object is constructed by adding the columns of $\Basis$, and that each column can be added only once.  This in turn means that the resulting structure estimate is constructed by adding components together.

In practice, the silhouettes of the various template shapes inevitably overlap.  The fact that these silhouettes are composited linearly in~\eqref{eq:Relaxed} means that 
pixels in overlapping areas of the reconstructed silhouette $\Basis \coeffs$ are no longer necessarily less than or equal to $1$.  In order to overcome this problem we apply a post estimation filtering step which is described in Section~\ref{sec:Randomized}.

%

\subsection{Dimension reduction by random projections}
\label{sec:Projections}


 The vector $\y$ in~\eqref{eq:Relaxed} is $\SilLength$ elements long, as are the columns of $\Basis$.  If $\y$ is constructed from $4$ images of one megapixel each then $\SilLength = 4\times 10^6$, and~\eqref{eq:Relaxed} will be too large to solve.  Given the sparsity referred to above, and the fact that we expect the columns of $\Basis$ represent a small fraction of the $2^\SilLength$ possible silhouettes, we randomly project the problem to a lower dimensional space and solve it there.

The Johnson-Lindenstrauss Lemma asserts~\cite{Arriaga99} that a set of $n$ points in any
Euclidean space can be mapped to an Euclidean space of dimension
$\Hashk = O(\epsilon^{-2}\mbox{log} \Hashn)$ so that all distances are preserved up to a multiplicative
factor between $(1-\epsilon)$ and $(1 + \epsilon)$. 
A variety of such mappings based on multiplication by a $\mathbb{R}^{\Hashk \times \Hashn}, \Hashk \ll \Hashn$ matrix $\Hash$ have been proposed, including that of Indyk and Motwani\shortcite{indyk1998approximate} where the elements of $\Hash$ are sampled independently from $\mathcal{N}(0,1)$, a zero-mean Gaussian distribution with standard deviation $1$.  Matou\v{s}ek in~\shortcite{Matousek08} showed that a sparse matrix with  nonzero entries chosen randomly from $\{1,-1\}$ can also be used.

In practice we apply dimension reduction by pre-multiplying the measurement and reconstruction by a matrix $\Hash \in \Rdim^{\nMeasurements \times \SilLength}$ where $\nMeasurements \ll \SilLength$ is chosen to ensure that the relative distances between each vector will be preserved with high probability.  The problem we wish to solve thus becomes
\begin{equation}
\argmin_\coeffs \|\Hash\y - \Hash\Basis \coeffs \|_1 + \lam \|\coeffs\|_1, \mbox{ s.t. } 0 \leq \coeffs \leq 1,
\label{eq:Hashed}
\end{equation}
The matrix $\Hash$ is constructed once, and is common to all templates.  The matrix is 
also 
 sparse, with only $\PhiDensity$ non-zero elements, and each such non-zero element drawn from a Normal distribution with mean $0$ and standard distribution $1$.  This form of $\Hash$ is a slight extension of the dense Gaussian matrix proposed by Motwani\shortcite{indyk1998approximate}, obtained by expanding the matrix with zeros.  The number of measurements required has not been changed, only the density with which they are represented in $\Hash$.
Various values of $\PhiDensity$ have been used in testing, as is detailed in Section~\ref{sec:Results}.

%
%
%

\subsection{Optimisation}
\label{sec:Practical}

A number of linear programming packages are capable of solving optimisation problems in the form of \eqref{eq:Hashed}, including CVX\footnote{ See http://www.cvxr.com.}
$\lone$-magic\footnote{See http://users.ece.gatech.edu/~justin/l1magic/.}, and MOSEK\footnote{ See http://www.mosek.com/..}, amongst many others.  

The columns of the matrix $\Basis$ are of the order of millions of elements long, and thus difficult to store and process.  Rather than store $\Basis$ itself, we thus record instead the $\Rdim^{\nMeasurements \times \nTemplates}$ basis matrix $\HBasis$ constructed by multiplying 
the basis matrix 
by the dimension reduction matrix thus
\begin{equation}
	\HBasis = \Hash \Basis.
\label{eq:Hasis}
\end{equation}
This projection is carried out as soon as 
each column of $\Basis$
is calculated, alleviating the need to store $\Basis$.

\begin{algorithm}
\caption{Algorithm overview}
\label{alg:overview}
\begin{algorithmic}
\REQUIRE Input images, Matrices $\Hash$, $\HBasis$
\STATE Calculate target object silhouette, and vectorise 
\STATE Eliminate columns of $\HBasis$ falling outside target silhouette
\STATE Solve Equation~\eqref{eq:Hashed} for $\coeffs$
\STATE Round $\coeffs$ to an integer solution 
\end{algorithmic}
\end{algorithm}

The time required to find the optimum of~\eqref{eq:Hashed} depends largely upon $\nTemplates$ the number of templates used and $\nMeasurements$, the number of rows in $\Hash$.  Decreasing $\nMeasurements$ decreases the probability that the optimum of~\eqref{eq:Hashed} will correspond to that of~\eqref{eq:Relaxed}, but culling unnecessary templates has no impact upon the quality of the solution.  
In practice it is often possible to eliminate templates from consideration by analysis of the images of the object to be reconstructed.  The simplest culling is achieved by removing from consideration those templates with silhouettes extending significantly beyond that of the real object.  This typically achieves an order of magnitude reduction in the number of columns in $\Basis$ in our testing.  

\subsection{Obtaining an integer solution}
\label{sec:Randomized}

The estimate $\est{\coeffs}$ generated through the process described above has elements which are bounded by $0$ and $1$, but which are not necessarily integers.  The cost function which is optimised by the linear programming process is also an approximation to the ideal cost function due to the linear composition process.
%

In order to generate an integer solution, and overcome the composition problem, a search method which checks every feasible solution against an improved evaluation function has been developed.  
It is typically possible to determine a range of numbers of templates that might be expected in a solution.  In the Lego example, for instance, the models might be expected to have between 5 and 20 parts.
A set of feasible templates can also be estimated by selecting every element of $\est{\coeffs}$ over a threshold.  Every possible combination of feasible templates whose cardinality is within a pre-defined range is thus determined.  This large set of possible template combinations is then sorted by the mean of the associated template coefficients.  The template combinations corresponding to the highest means are then evaluated. 

Evaluating a template combination involves rendering its silhouette and comparing it to the true silhouette of the object. The silhouette of each template combination is calculated so as to avoid the linear composition problem, and thus allows a true evaluation of the quality of each combination.  The combination with the lowest silhouette reconstruction error, minus lambda times the number of templates, is selected as the final result.
We call this the `Search' method, as compared to the `Max' method which merely selects a fixed number of templates corresponding to the highest entries of $\coeffs$.  The Max method requires knowledge of the true number of templates used, whereas the Search method does not, but requires more computation to evaluate candidate template combinations.


\section{Experimental testing}
\label{sec:Results}

We test the accuracy of the recovered model structure, and its sensitivity to noise, for both synthetic images and images of real objects. Additionally, we measure the effect of varying parameters such as the number of shape templates and the degree of dimension reduction.

\begin{figure}[htb]
	\centering
		\includegraphics[width=0.29\columnwidth]{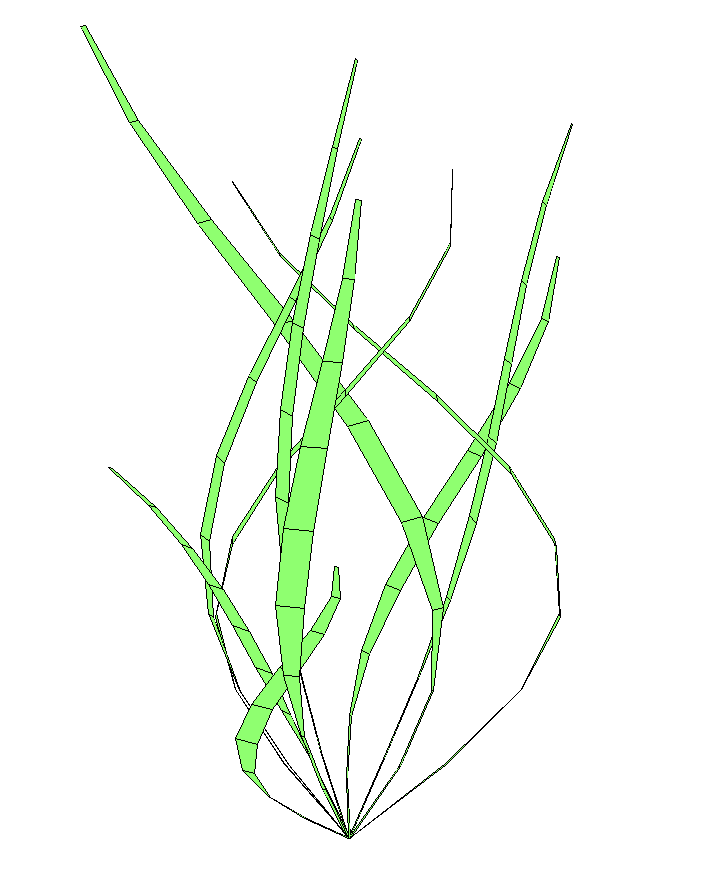}
		\includegraphics[width=0.24\columnwidth]{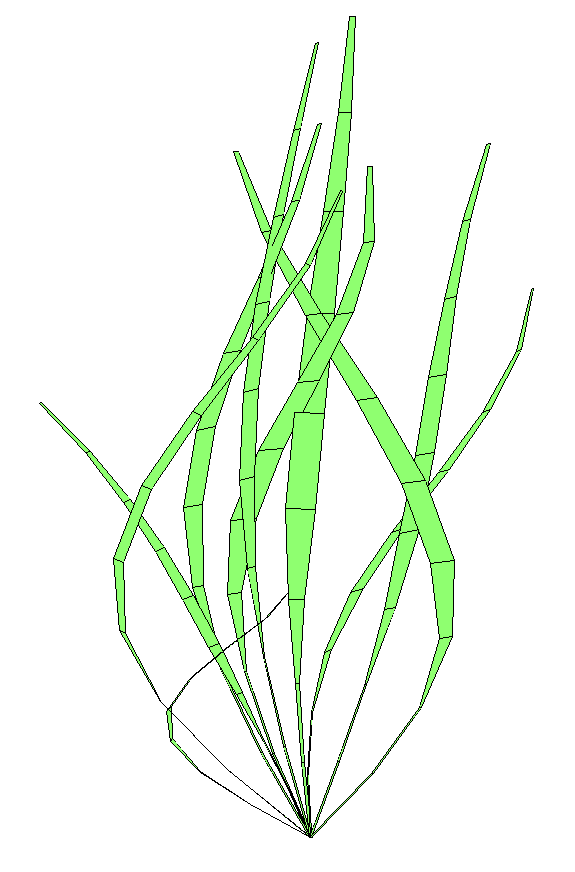}
		\includegraphics[width=0.4\columnwidth]{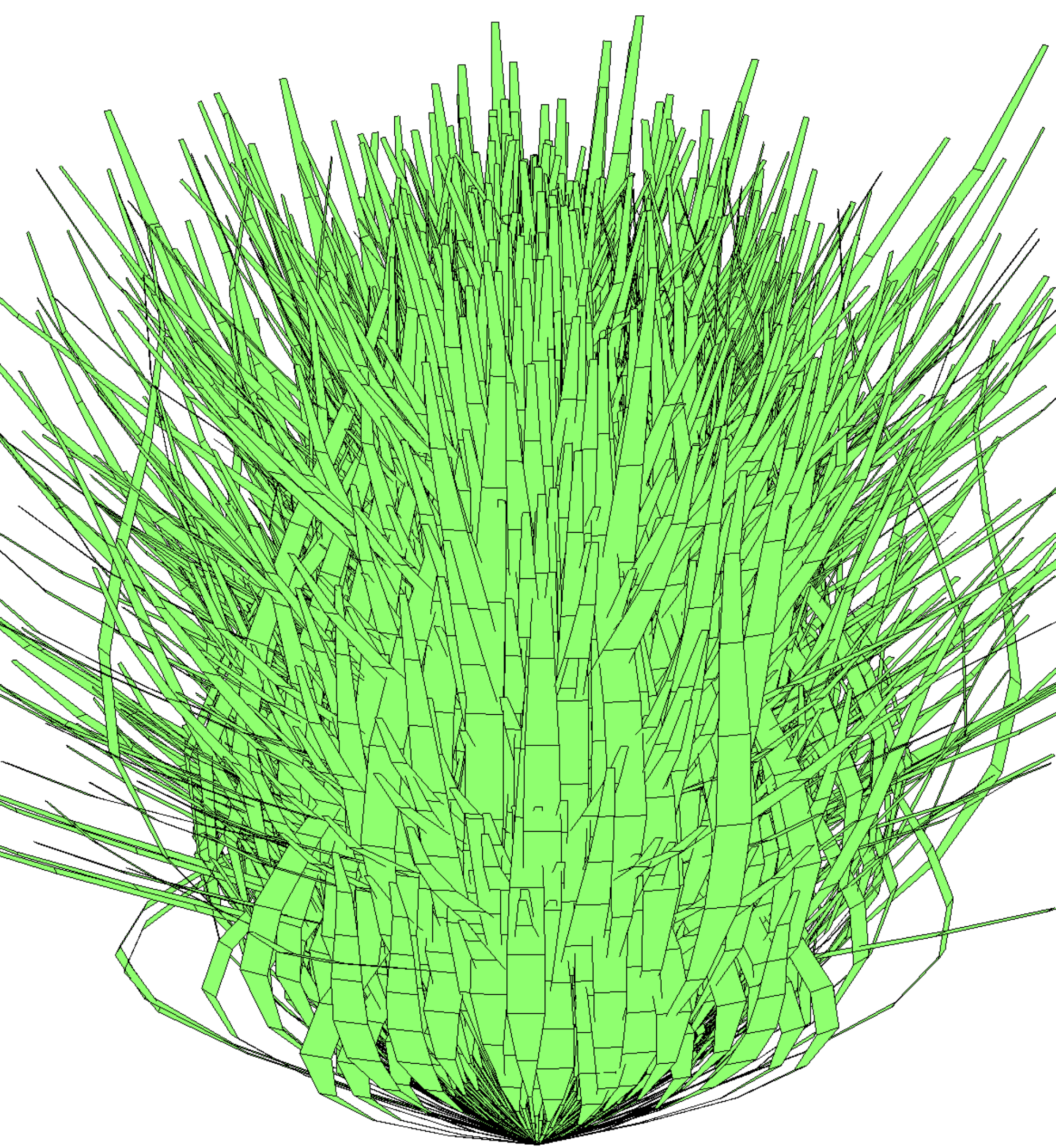}
	\caption{Two views of a synthetically rendered plant model used in testing, and a rendering of $2000$ leaves which illustrates the amount of overlap between templates. }
	\label{fig:SynthPlant}
\end{figure}

The model used for synthetic testing was that of a hypothetical plant composed of single leaves emanating from a common root, as illustrated in Figure~\ref{fig:SynthPlant}.  The testing, and rendering, was carried out in Matlab, allowing full control of plant parameters.  Optimisation was carried out using the CVX package\footnote{See http://cvxr.com/cvx/}.  Unless otherwise specified, synthetic tests on the plant model used four images (and thus silhouettes) per plant, each of $281 \times 211$ pixels, the plants had 6 leaves, and each result represents an average over 10 trials.  The default parameters to the method were $\lam = 10^{-2}$, $\nMeasurements = 441$, $\PhiDensity = 10^{-2}$  and $\nTemplates = 500$.  The values of these parameters were determined through training on separately generated test data.  


\begin{figure}[htb]
	\centering
		\includegraphics[width=0.49\columnwidth]{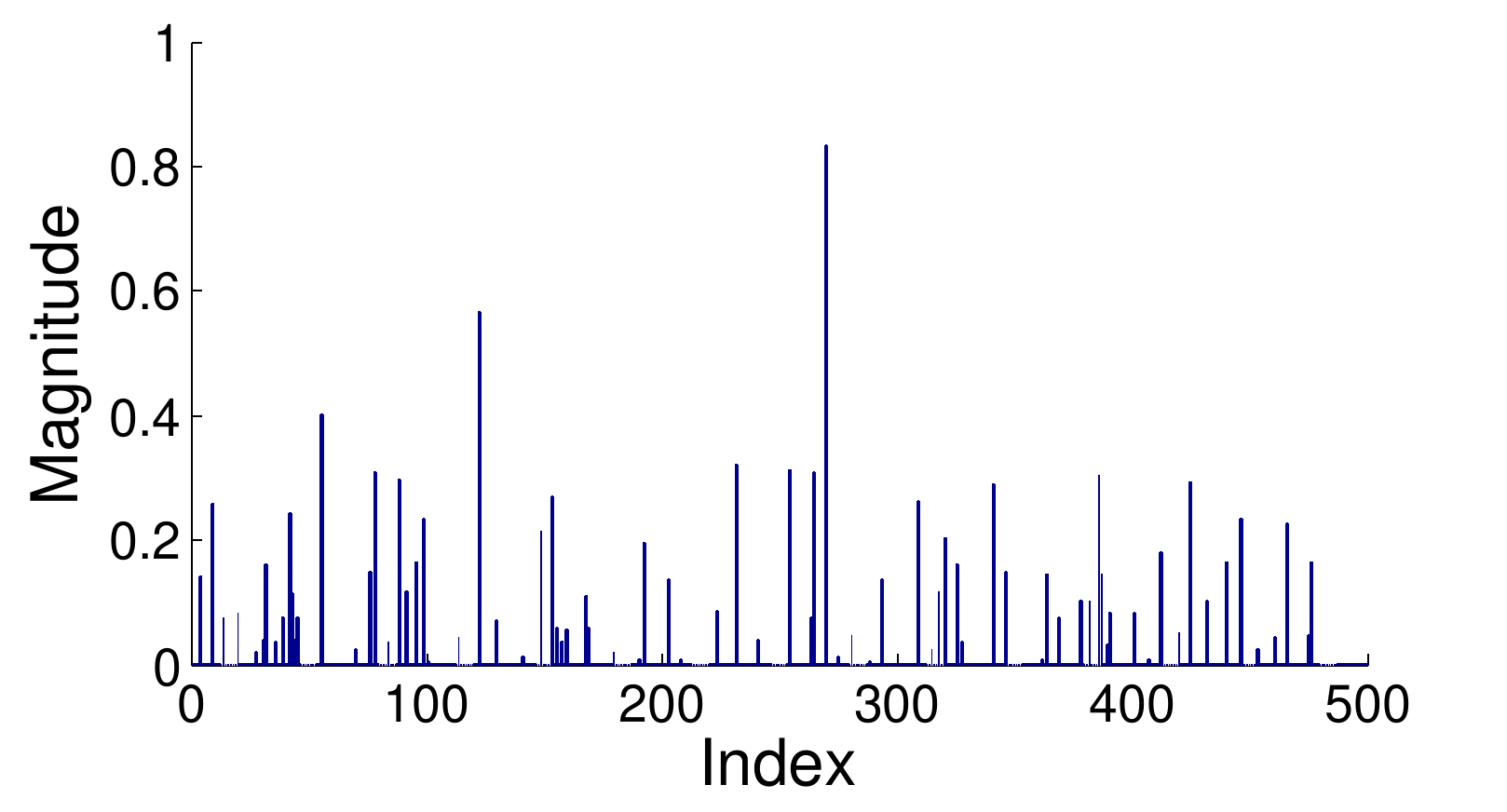}
		\includegraphics[width=0.45\columnwidth]{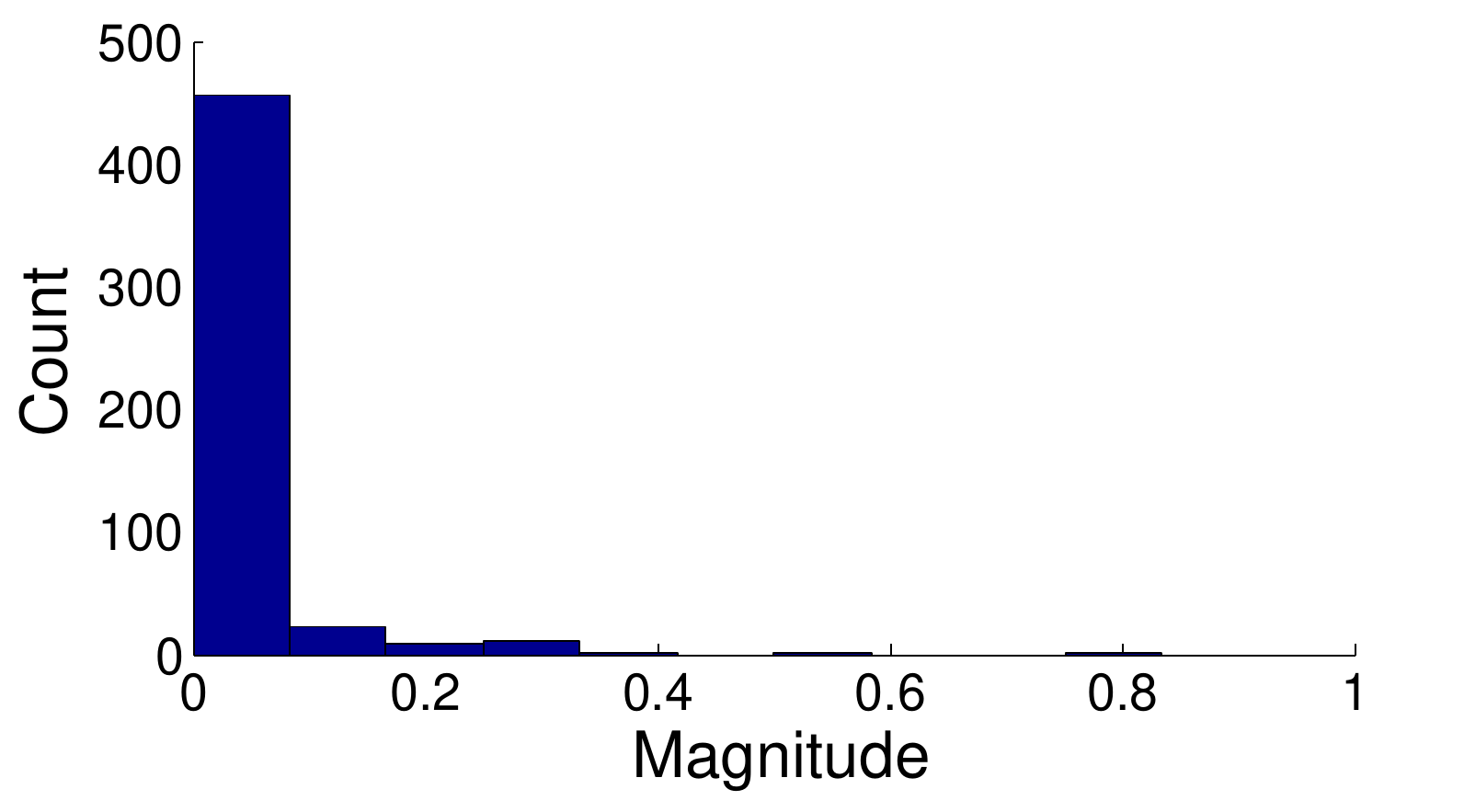}
	\caption{A bar and histogram plot of an $\coeffs$ estimated by the optimisation process. The true model had 16 components, and a significant amount of noise added to the model parameters (the fraction of pixels changed was $0.075$).}
	\label{fig:AlphaBarPlot}
\end{figure}

Figure~\ref{fig:AlphaBarPlot} shows an $\coeffs$ estimated by the linear programming process, and a histogram of the magnitudes of its entries.  Note that the recovered $\alpha$ is very sparse, but that the number of coefficients greater than zero is significantly larger than the true number of leaves, which was $16$.

We measure performance by the proportion of pixels in the test silhouettes which are explained by the estimated structure.  This is calculated by building the model corresponding to the estimated $\coeffs$ and comparing its silhouette to the test silhouette.  This error metric has the advantage of being relatively independent of the choice of the basis $\HBasis$ because, as is stated above, we seek the structure estimate which best explains the silhouette.  We label this measure the `Fraction of Pixels Explained' (FPE), and the ideal result is that the fraction of pixels explained by the estimate is equal to the fraction of pixels explained by the true structure.  Such a result would fall on the line $y = x$ within the plots shown below of FPE against Noise Level (Fraction of Pixels Changed).

\begin{figure}[htb]
	\centering
		\includegraphics[width=\columnwidth]{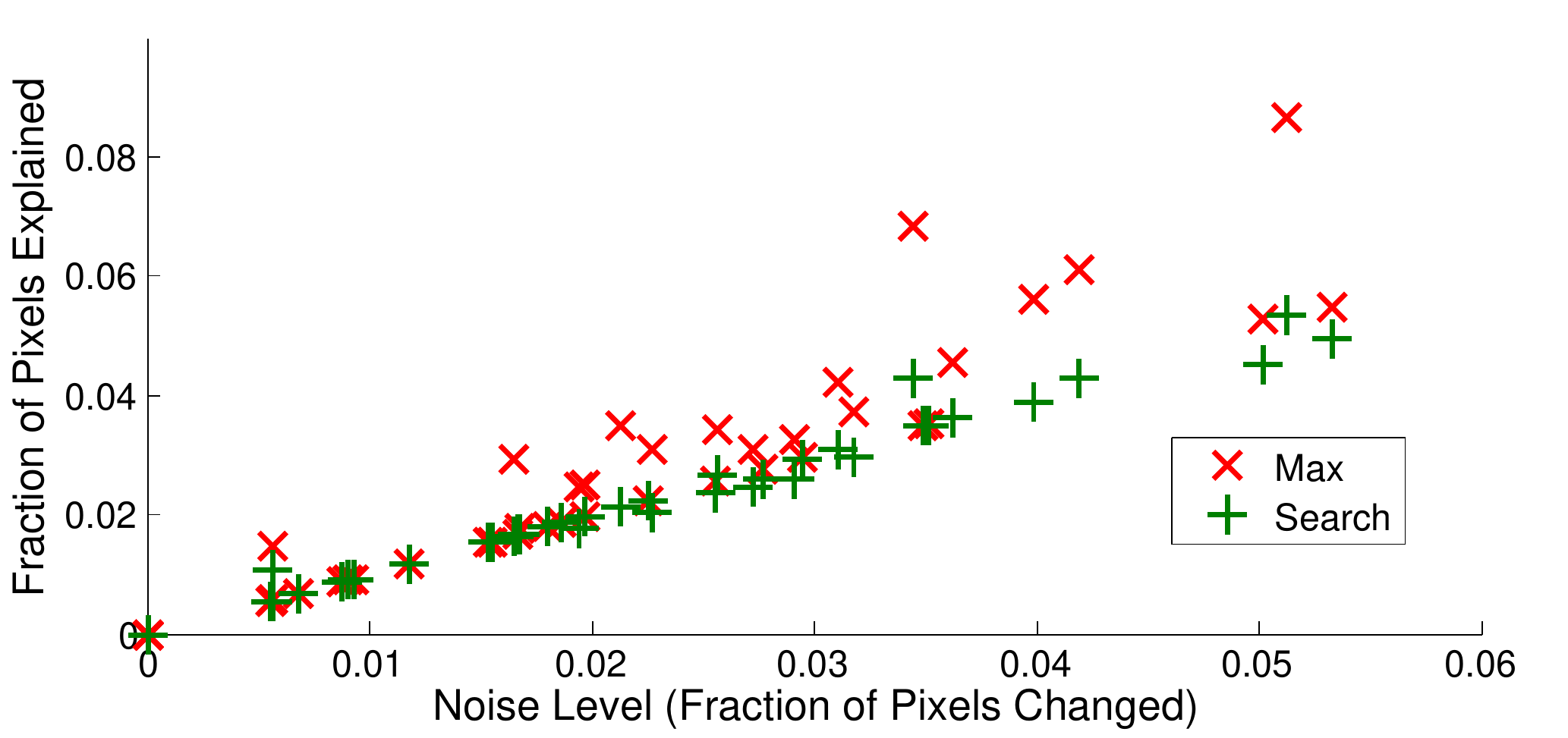}

	\caption{A scatter plot of Noise level vs. FPE for structure estimates calculated by the `Max' method (red crosses) and the 'Search' method (green plus symbols).}
	\label{fig:ScatterPlot-BestVsMax-2NoiseLevels-6and8Leaves}
\end{figure}

Figure~\ref{fig:ScatterPlot-BestVsMax-2NoiseLevels-6and8Leaves} illustrates the performance of the method over $200$ tests with varying levels of parameter noise, and numbers of leaves.  The fact that the results of the `Search' process (from Section~\ref{sec:Randomized}) are closely aligned to the line $y = x$ indicates that the method is calculating estimates which explain the observations as well as possible.

\begin{figure}[htb]
	\centering
		\includegraphics[width=\widthfrac\columnwidth]{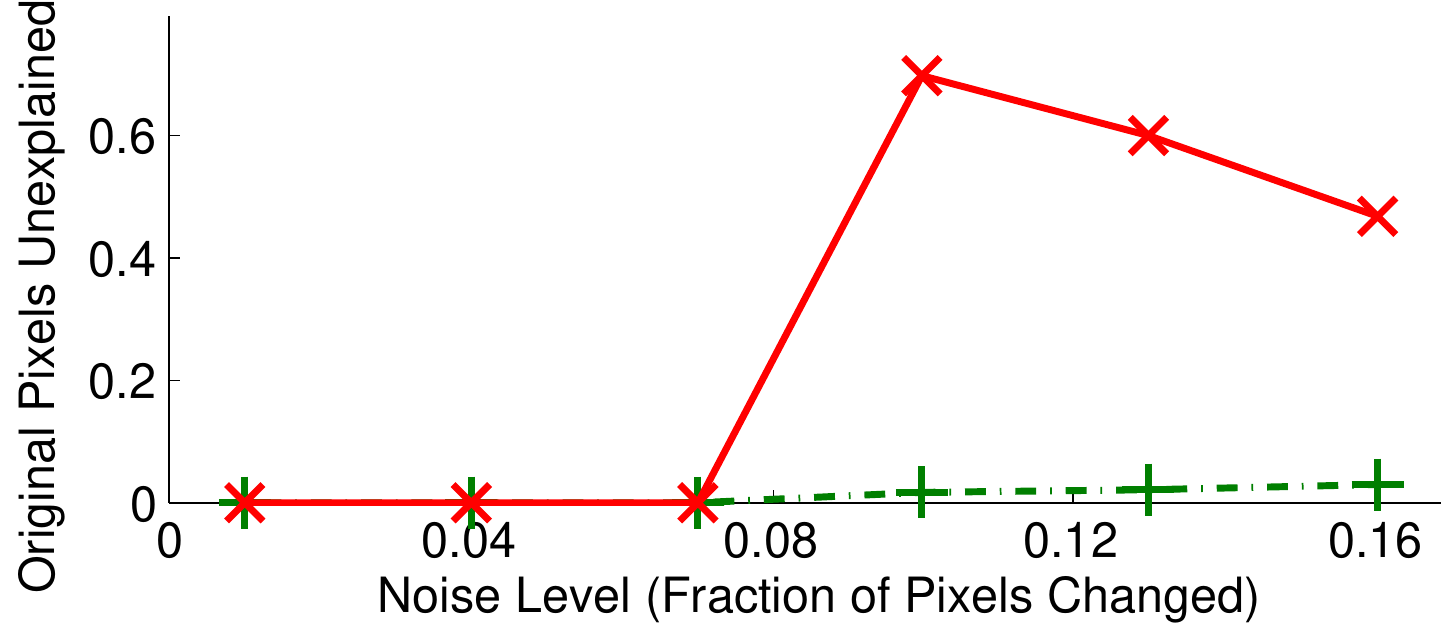}
	\caption{The performance of the `Max' and `Search' methods (in red and blue respectively) when applied to silhouettes corrupted with salt and pepper noise.  The $y$-axis relates the number of pixels in the noise free image that are not explained by the reconstruction, and the $x$-axis the fraction of pixels that had their values set randomly.}
	\label{fig:SAndPOriginalPixelsUnexplained}
\end{figure}

Figure~\ref{fig:SAndPOriginalPixelsUnexplained} relates the performance of the method in the presence of salt and pepper pixel noise.  In generating the figure a fraction of the pixels in each silhouette were set randomly to zero or one.  The Figure shows that the method is very robust to this type of error, with the `Search' method still performing well in the case where $16\%$ of the pixels have been changed. This is to be expected due to the use of templates as the structural elements rather than individual points or voxels.

\begin{figure}[htb]
	\centering
		\includegraphics[width=\widthfrac\columnwidth]{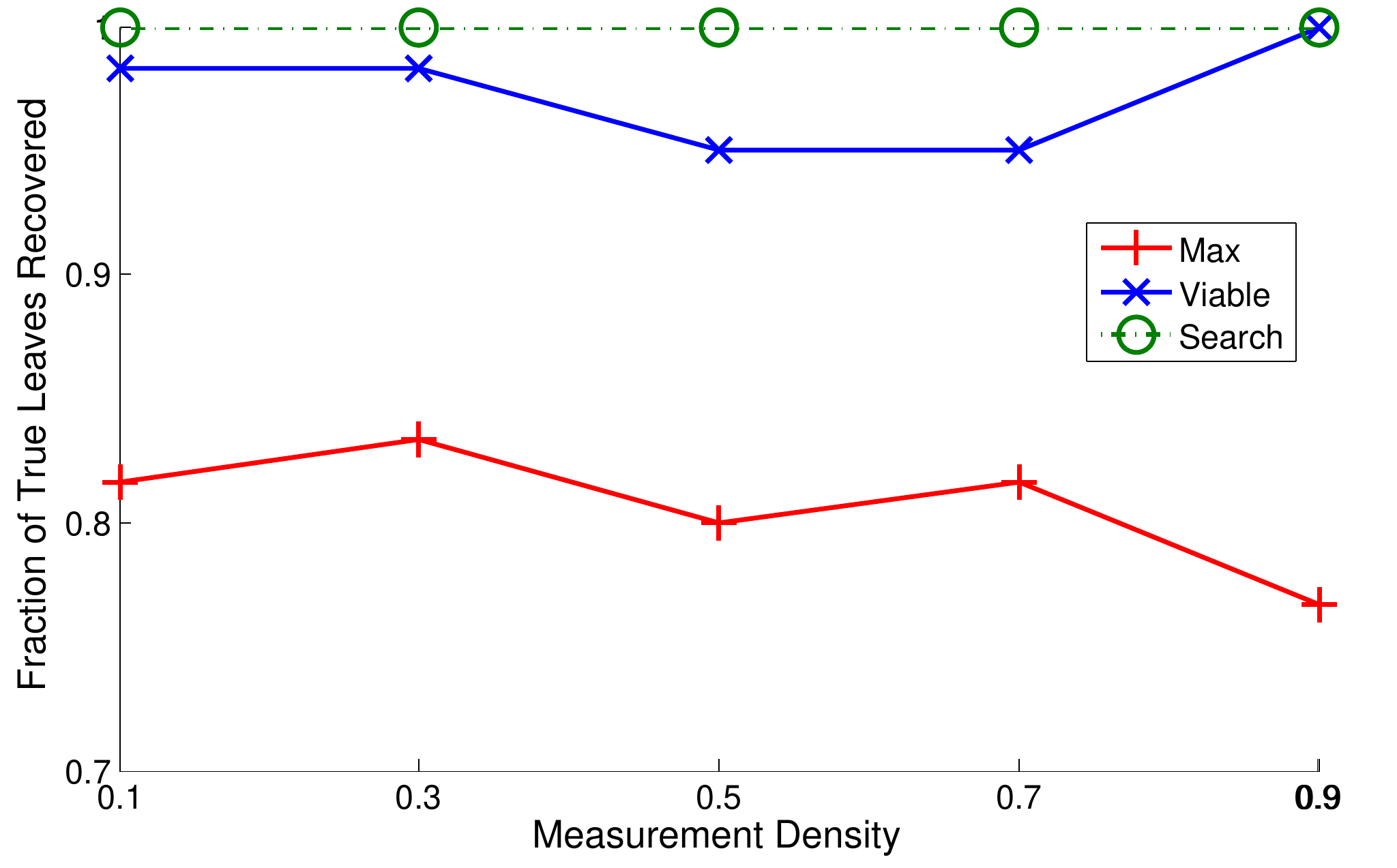}
	\caption{The fraction of true leaves recovered for varying measurement densities $\PhiDensity$.   
	}
	\label{fig:MeasurementDensityPlot-2}
\end{figure}

\begin{figure}[htb]
	\centering
		\includegraphics[width=\widthfrac\columnwidth]{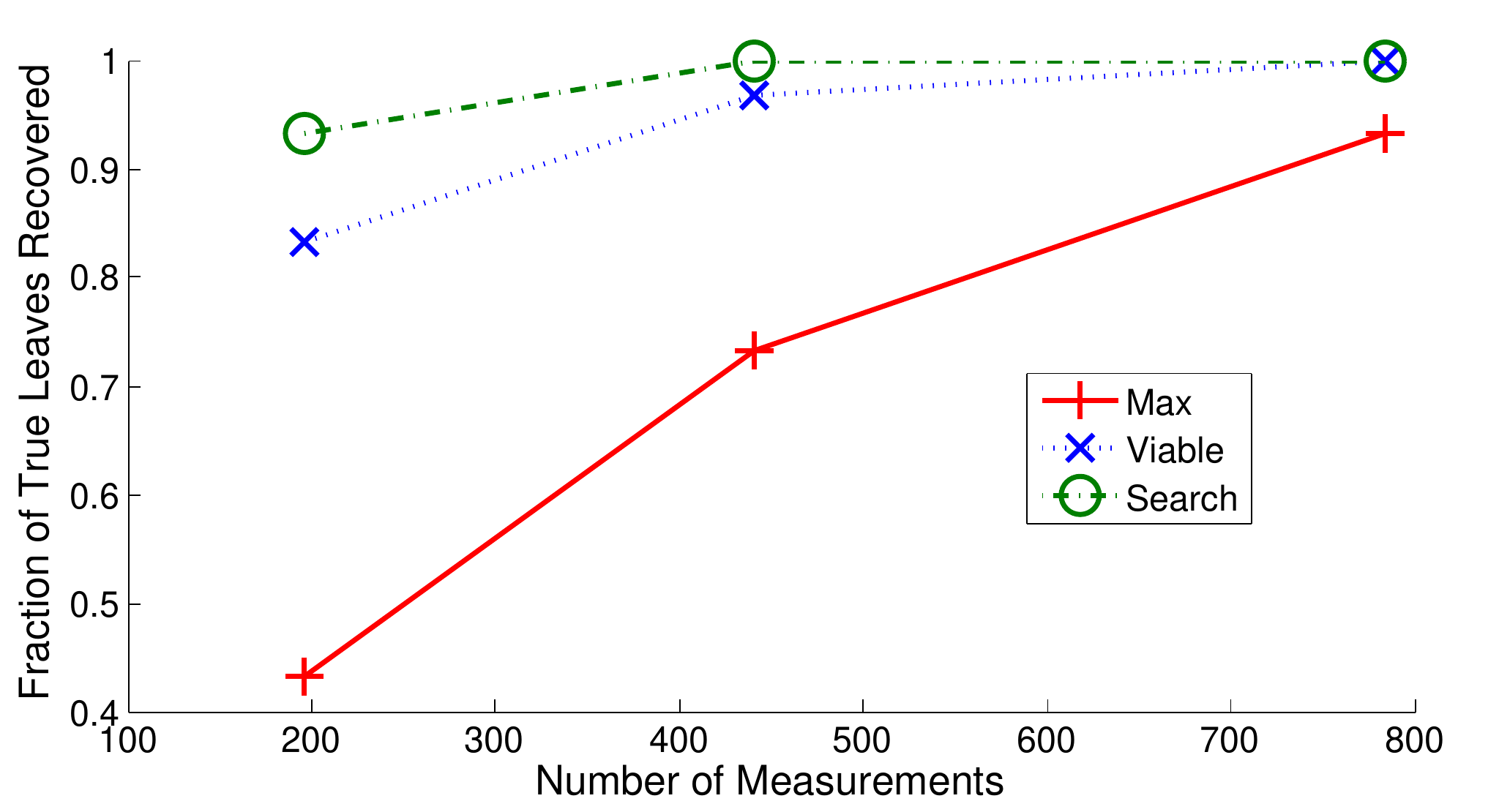}
	\caption{The fraction of true leaves recovered for varying numbers of measurements (or projection dimensions) $\nMeasurements$.}
	\label{fig:VaryingNumberOfMeasurements}
\end{figure}

Figure~\ref{fig:MeasurementDensityPlot-2} shows the proportion of plant leaves recovered under various values for the measurement density $\PhiDensity$. The tests were carried out in the presence of salt and pepper noise whereby $7\%$ of the pixels in each silhouette had their values assigned randomly.  The figure shows that the Search method is robust to changes in measurement density. As expected, the accuracy of the reconstruction improves as the number of measurements increases in Figure~\ref{fig:VaryingNumberOfMeasurements}.

\subsection{Real images tests}
\label{sec:RealImages}


Figures~\ref{fig:teaser} and \ref{fig:lego_teaser} relate the performance of the method in estimating structure from real images.
In the real image testing the pictures were taken with a standard consumer SLR camera and 
the silhouettes
calculated using standard image processing techniques.  Forming the matrix $\HBasis$ required of the order of $4$ minutes which includes rendering time.  The Lego shape templates were generated using LeoCAD\footnote{See 
http://www.leocad.org/trac} which in turn uses the models from LDraw.org\footnote{See http://www.ldraw.org/}.
The templates represent different blocks at all possible positions and rotations within the bounds of the space within which the model may appear.  In order to restrict the number of templates to a manageable number only those with silhouettes which overlap the real object silhouettes are considered. For all real image tests $\lam = 4\times 10^{-4}$.


\begin{figure}[htb]
	\centering
		\includegraphics[width=\columnwidth]{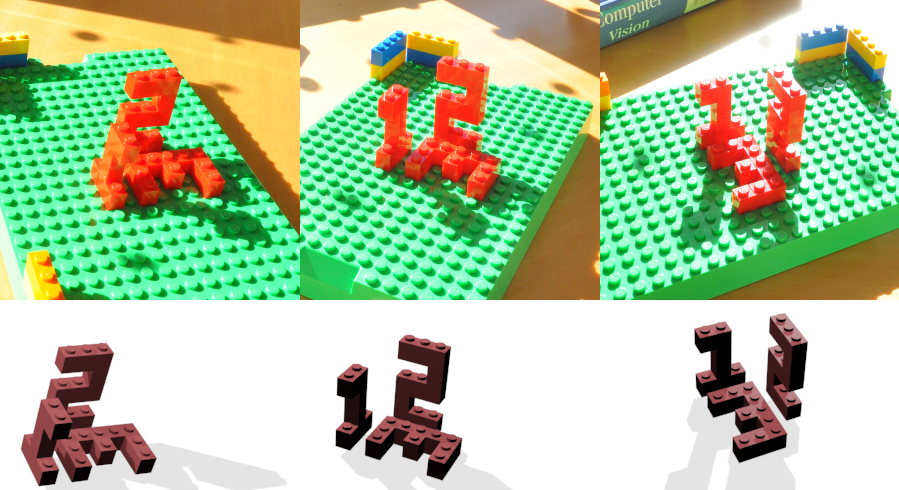}
	\caption{Three of the 8 input images for the numbers test, and the corresponding views of the recovered model.
	The only shape model used in the test was a $1 \times 1$ Lego block.}
	\label{fig:Numbers}
\end{figure}

In the case of Figure~\ref{fig:lego_teaser} two 3D shape models and 4 images were used.  The images were calibrated using standard software on the basis of the locations of the corners of blocks in the Lego base.  In terms of templates, the $1 \times 1$ block could appear in $196$ positions horizontally and was modelled up to 4 layers high, resulting in $784$ templates.  The arch had $121$ possible positions (horizontally) at each of 2 orientations, which at $4$ level generated $968$ templates.  The total problem thus had $1,752$ templates, which was reduced to just over $500$ after culling.  The number of measurements $\nMeasurements$ used was $1,727$
and solving the system required approximately 3 minutes of processing time.  The timing information is approximate as much of the process was parallelised and calculating the equivalent serial timing necessarily involves estimation.

\begin{figure}[htb]
	\centering
		\includegraphics[width=0.9\columnwidth]{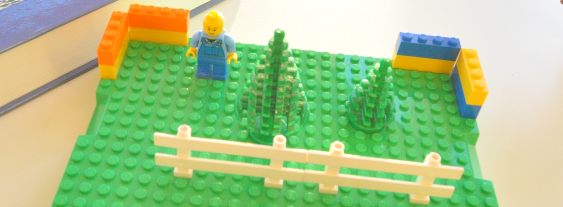}
		\includegraphics[width=\columnwidth]{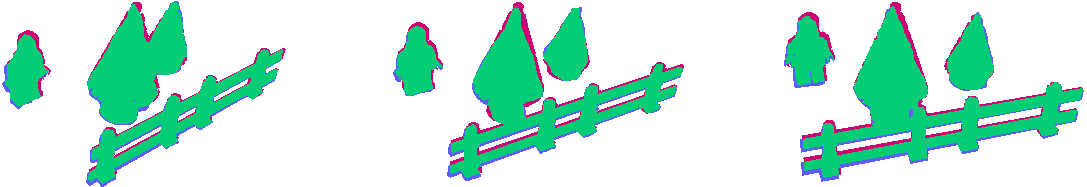}
		\caption{A segment from one of 8 input images for the trees test, and a visualisation of the reconstruction projected back into the first 3 input images.  The projection of the reconstruction is shown in red, and the original silhouette in blue.  Green thus indicates overlap.  The results are effectively the same for $\lam \in [1\times 10^{-5},6\times 10^{-3}]$.  Nine shape models were included in the test. The fact that the reprojection and silhouette are slightly misaligned demonstrates the robustness of the method to errors in calibration and segmentation.}
	\label{fig:Trees}
\end{figure}

In the case of Figure~\ref{fig:teaser} four 3D shape models were used, the figure, the windscreen, the wheel pairs, and a $6 \times 2$ block.  The number of templates after culling was $861$.  Rendering, segmenting and projecting the templates took approximately 6 minutes, solving the system approximately 4 minutes, and the search process 9 minutes.

\begin{figure}[htb]
	\centering
		\includegraphics[height=2.5cm]{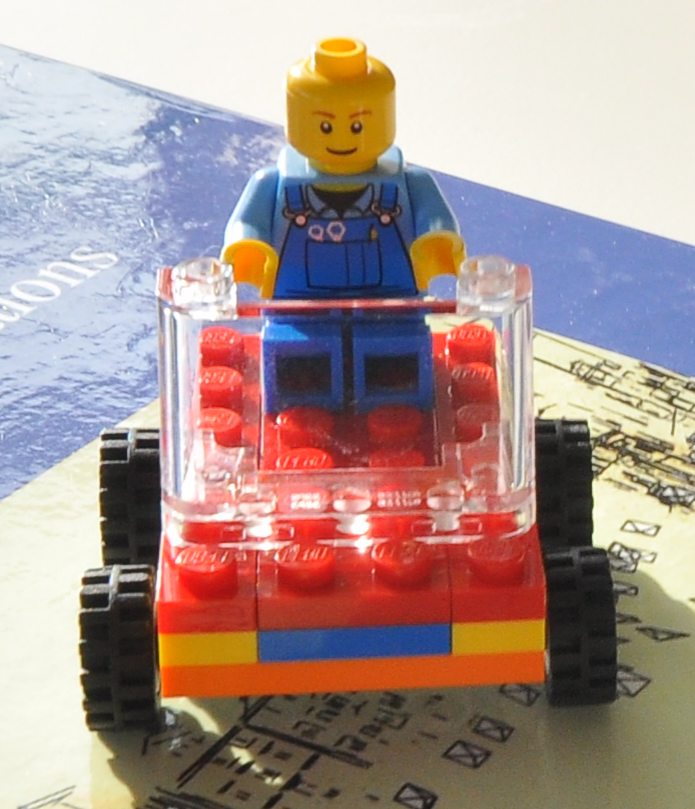}
		\includegraphics[height=2.5cm]{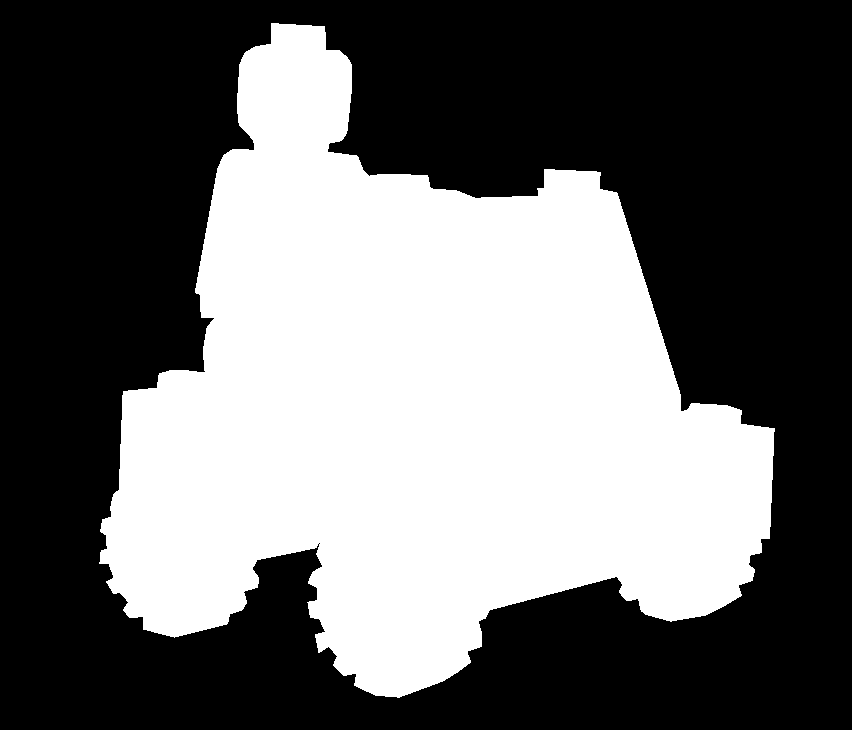}
		\includegraphics[height=2.5cm]{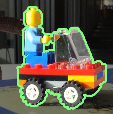}
		\includegraphics[width=\columnwidth]{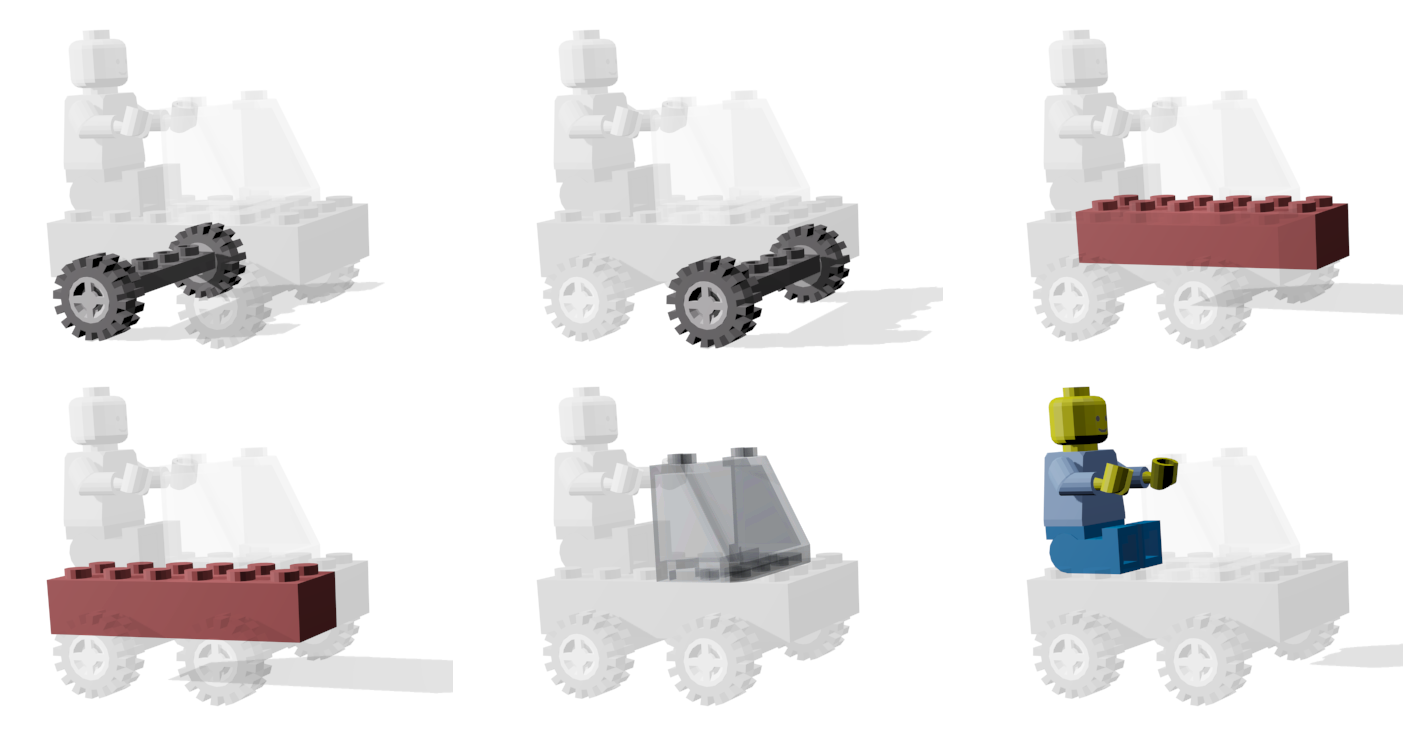}
	\caption{One of 12 input images for the second car test (cropped), one of the calculated silhouettes, the reprojection of the ground truth back into one of the images, and rendering of the estimated structure.
	Three times as many shape models as required were used in the test (shown in Figure~\ref{fig:LegoPieces}), leading to over $3,500$ templates.  The result is stable for $\lam \in [0,2\times 10^{-3}]$.}
	\label{fig:car2}
\end{figure}

\begin{figure}[htb]
	\centering
		\includegraphics[height=3.5cm]{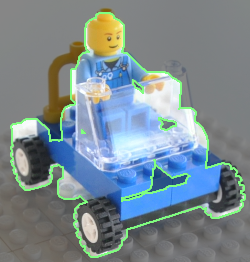}
		\includegraphics[height=3.5cm]{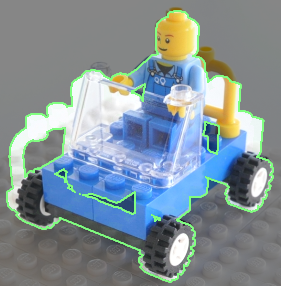}
	\caption{Failure case. Two of the input images with the silhouette of the estimated reconstruction to illustrate the discrepancy between the two.}
	\label{fig:RealSilhouettes}
\end{figure}

There are a number of failure cases of the method.  The correct calibration of the image set is critical, as even small errors will see the synthetic and real components of the object misaligned.  
The failure case presented in Figure~\ref{fig:RealSilhouettes} illustrates the case in which the overlap between template silhouettes grows to be a significant proportion of the silhouette itself.  The templates used in deconstructing the model included $1 \times 1$ blocks.  These blocks have a very small projection into the images, and a large proportion of each is occluded by the blocks in front of it.  The size of the blocks also means that many templates are required to enumerate the space.  The result is a large number of templates which have a minimal impact on the object silhouette.

\section{Conclusion}


The method we have described recovers a structural explanation of the shape of an object from a set of silhouettes.  It is applicable in cases where structure can be described in terms of a set of building blocks, and where the this set 
is of the order of millions of elements or less.  Experimental testing has shown that it is robust to salt and pepper noise, and to moderate discrepancies between real object structures and the set of template shapes.

Estimating object structure from an image set is often under-constrained, as many possible structures could generate the same set of observations.  This is typically the case where an object includes internal components invisible in the images, for instance.  
In this case the method choses the structure with smallest number of elements, but other regularisations are equally feasible.

The primary limitation of the method is the fact that the linear composition model from Equation~\eqref{eq:1} generates silhouette reconstructions with values greater than $1$.  This limits the complexity of the objects which may be deconstructed, as complex objects tend to have higher levels of overlap in their components.  We have demonstrated that the method is capable of selecting $20$ components from a set of over $10,000$, but this will need to be improved for many real applications.
Lessening this limitation remains as further work, as does the development of template learning methods, and possible the use of non-silhouette features.  

One of the most interesting features of the method is that it optimizes over combinations of 3D building blocks, and to this extent is reasoning in 3D about the problem of understanding objects from image sets.  An interesting  extension would be the inclusion of physical constraints in the optimisation process,
 such as those utilised in~\cite{gupta10blocks}.  These might include the requirement that all estimated components are subject to gravity, for example.

The proposed approach has been demonstrated to succeed for compound objects of moderate complexity, and offers insight into the problem and a basis upon which further developments might proceed.  

%

{\small
\bibliographystyle{ieee}
\bibliography{Plants}
}

\end{document}